\documentclass{article}
\usepackage{spconf,amsmath,graphicx}

\usepackage[table]{xcolor}
\usepackage[dvipsnames]{xcolor}
\usepackage{subcaption}
\usepackage{enumitem}
\usepackage{verbatim}
\usepackage{amsthm}

\usepackage{wrapfig}
\usepackage{subcaption}
\usepackage{float}

\usepackage[utf8]{inputenc} % allow utf-8 input
\usepackage[T1]{fontenc}    % use 8-bit T1 fonts
\usepackage{url}            % simple URL typesetting
\usepackage{booktabs}       % professional-quality tables
\usepackage{amsfonts}       % blackboard math symbols
\usepackage{nicefrac}       % compact symbols for 1/2, etc.
\usepackage{microtype}      % microtypography
\usepackage{xcolor}         % colors
\usepackage{multirow}
\usepackage{graphicx}
\usepackage{amsmath}
\usepackage{bbding}

\usepackage[colorlinks=true, filecolor=blue, linkcolor=blue, urlcolor=blue, citecolor=blue]{hyperref}
\usepackage{cleveref} 
\usepackage{subcaption}

\usepackage{graphicx}
\usepackage{tikz}
\usetikzlibrary{arrows.meta,positioning}
\title{Training Crossroads for Recurrent Vision Transformers: \\ Recurrence, Neural ODEs, and Deep Supervision
}
\name{Grzegorz Gruszczynski$^{1*}$\thanks{*Corresponding author. e-mail:  g.gruszczyns@samsung.com} \quad Pawel Olszowiec$^{1}$ \quad Michal Byra$^{1,2}$  
\quad Grzegorz Stefanski$^{1}$  \quad Alberto Presta$^{1}$ }
\address{$^{1}$Samsung AI Center, Warsaw, Poland  \quad \\ $^{2}$Institute of Fundamental Technological Research, Polish Academy of Sciences, Warsaw, Poland}
\begin{document}
%\ninept
%
\maketitle
\begin{abstract}
Vision Transformers (ViTs) achieve strong image-recognition performance,
but their parameter count grows linearly with depth when each block is
independently parameterized. Single-block recurrent ViTs (bViT) remove
this growth by repeatedly applying one shared block.
Rather than proposing a new architecture, we fix a bViT and provide a controlled
empirical characterization of three training and inference regimes under
a common CIFAR-100 protocol, asking: (i)~when does recurrence beat
independently parameterized depth---at matched FLOPs or at matched
parameter memory? (ii)~when a residual recurrent block is trained
\emph{through} an ODE solver, does solver order act as numerical
refinement or as an architectural bias? and (iii)~what does robustness
beyond the training horizon cost in nominal accuracy? We find that
standard ViTs remain preferable when FLOPs are the primary constraint,
whereas recurrent ViTs offer a better accuracy--parameter trade-off
under memory constraints. Consistent with the standard view of residual
networks as Euler discretizations of ODEs, the continuous-time analogue
of a residual recurrent block is the state-subtracted vector field
$\dot{z}=F_\theta(z)-z$; although known in principle, 
this distinction is easy to violate when the block is wrapped as a black-box vector field, and we qualify the cost at few accuracy points. 
Because the vector field is learned jointly with the solver, higher-order solvers act as a solver-induced architectural bias rather than a numerical-accuracy improvement, 
and their gains are not uniform.
Finally, stage-wise deep supervision traces an accuracy--robustness frontier: it does not improve nominal accuracy, but degrades gracefully far beyond the training horizon, where naive recurrence collapses to near-random performance.
\end{abstract}


\section{Introduction and Related Work}

Vision Transformers (ViTs) have established \textit{state-of-the-art} performance in image recognition but incur high memory costs due to their deep, 
independently parameterized block structures \cite{caron2021emerging,dosovitskiy2020image,he2022masked,touvron2021training}. 
While empirical evidence suggests that transformer layers often perform highly related computations and can be compressed or merged \cite{akiba2025evolutionary,garg2025revealing,sun2025transformer}, 
standard architectures still scale their parameter footprint linearly with depth. 
Recurrence and parameter sharing offer a mechanism to decouple computational depth from parameter count, 
predominantly in language modeling, synthetic reasoning, or through partial sharing, channel slicing and distillation in vision \cite{dehghani2018universal,jacobs2025block,lan2019albert,saunshi2025reasoning,zhang2022minivit,shen2022sliced}. 
In contrast to these  designs, the single-block Vision Transformer (bViT) investigates the extreme limit of full parameter sharing, 
replacing the entire sequence of unique layers with a single, recurrently applied operator trained from scratch \cite{byra2026bvitinvestigatingsingleblockrecurrence}.

\vspace{-0.825em}
\section{Architecture and Training Procedure}
\label{sec:model-architecture}
\begin{figure}[t!]
\centering
\resizebox{\columnwidth}{!}{%
\begin{tikzpicture}[
    font=\sffamily, 
    block/.style={
        draw,
        rounded corners=6pt,
        very thick,
        minimum width=2.65cm,
        minimum height=1.75cm,
        font=\sffamily\Large,
        align=center
    },
    arrow/.style={-Latex, very thick},
    dots/.style={font=\Large\bfseries},
    title/.style={font=\huge},
    steplabel/.style={font=\large, text=blue!45!black},
    note/.style={font=\huge}
]

% ------------------------------------------------------------
% Colors
% ------------------------------------------------------------
\definecolor{patchgray}{RGB}{210,207,207}
\definecolor{vitblock}{RGB}{255,192,0}
\definecolor{shared}{RGB}{255, 239, 125}
\definecolor{classifier}{RGB}{157,195,230}

\definecolor{vitgreenone}{RGB}{198,239,206}
\definecolor{vitgreentwo}{RGB}{146,208,80}
\definecolor{vitgreenthree}{RGB}{112,173,71}

% ------------------------------------------------------------
% Top: standard ViT
% ------------------------------------------------------------
\node[title] at (7.5, 2.65) {Standard ViT};

\node[block, fill=patchgray] (pe1) at (0,1) {Img\\Embed};
% \node[block, fill=vitblock]  (b1)  at (3.2,1) {Block $1$};
% \node[block, fill=vitblock]  (b2)  at (6.4,1) {Block $2$};
\node[block, fill=vitgreenone]   (b1)  at (3.2,1) {Block $1$};
\node[block, fill=vitgreentwo]   (b2)  at (6.4,1) {Block $2$};
\node[dots]                  (d1)  at (9.0,1) {$\cdots$};
% \node[block, fill=vitblock]  (bL)  at (11.8,1) {Block $N$};
\node[block, fill=vitgreenthree] (bL)  at (11.8,1) {Block $N$};
\node[block, fill=classifier] (cl1) at (15.0,1) {Classifier};

\draw[arrow] (pe1) -- (b1);
\draw[arrow] (b1) -- (b2);
\draw[arrow] (b2) -- (d1);
\draw[arrow] (d1) -- (bL);
\draw[arrow] (bL) -- (cl1);

% ------------------------------------------------------------
% Separator
% ------------------------------------------------------------
\draw[very thick, dashed] (-1.4,-0.55) -- (16.4,-0.55);

% ------------------------------------------------------------
% Bottom: recurrent ViT
% ------------------------------------------------------------
\node[title] at (7.5,-1.45) {Recurrent bViT};

\node[block, fill=patchgray] (pe2) at (0,-3) {Img\\Embed};
\node[block, fill=shared] (s1) at (3.2,-3) {Shared\\Block};
\node[block, fill=shared] (s2) at (6.4,-3) {Shared\\Block};
\node[dots]                  (d2) at (9.0,-3) {$\cdots$};
\node[block, fill=shared] (sT) at (11.8,-3) {Shared\\Block};
\node[block, fill=classifier] (cl2) at (15.0,-3) {Classifier};

\draw[arrow] (pe2) -- (s1);
\draw[arrow] (s1) -- (s2);
\draw[arrow] (s2) -- (d2);
\draw[arrow] (d2) -- (sT);
\draw[arrow] (sT) -- (cl2);

\end{tikzpicture}%
}
\caption{Standard ViT (upper) vs.  bViT (lower) architecture.}
\label{fig:architecture}
\end{figure}
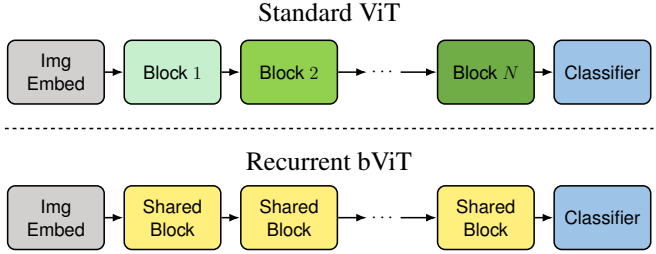

We fix the recurrent ViT architecture and dataset, and compare alternative training/inference paradigms.
% : finite-step recurrence, Neural ODE integration, and stage-wise deep supervision. 
The purpose of this study is not to establish a new SOTA architecture for image recognition task, 
but to identify how different computational training regimes affect accuracy, compute, parameter efficiency, solver behavior, and extrapolation stability.
We follow the standard Vision Transformer image-classification pipeline: an input image is split into non-overlapping patches, projected into token embeddings, pre-pended with a learnable \textsc{cls} token, and augmented with positional embeddings.
The resulting sequence is processed by pre-norm transformer blocks with multi-head self-attention and feed-forward sub-layers, and the final normalized \textsc{cls} token is passed to a linear classification head, see Fig. \ref{fig:architecture}. 

We adopt the single-block recurrent topology of Byra et al.~\cite{byra2026bvitinvestigatingsingleblockrecurrence}:
the stack of independently parameterized blocks $F_1,\ldots,F_L$ of a standard ViT is replaced by a single shared transformer block $F$ applied recurrently:
$z_{t+1} = F_\theta(z_{t})$ where $ t=0,\ldots,T-1$.
Figure~\ref{fig:architecture} shows that the bViT can be interpreted as an unrolled ViT with full parameter sharing across depth. 
By reusing the same multi-head self-attention and feed-forward components at every step, 
bViT substantially reduces the model's overall parameter count and memory footprint while preserving the standard computational budget.

\subsection{Evaluations}
We evaluate the models on an image classification task. 
We use CIFAR-100 \cite{krizhevsky2009learning} dataset, which contains 100 distinct classes of RGB images, each corresponding to 600 images, 500 and 100 for training and testing, respectively. 
Models are trained for 200 epochs over three independent runs. 
We report the mean best test accuracy across runs,
with per-configuration standard deviations ranging from $\pm0.001$ to
$\pm0.004$.

Training is performed using the AdamW optimizer \cite{loshchilov2017decoupled} with a learning rate of $1 \times 10^{-3}$ and a weight decay coefficient of 0.05 to minimize the standard cross-entropy loss. 
The default network and training hyperparameters are configured as follows~\cite{byra2026bvitinvestigatingsingleblockrecurrence}: 
an embedding dimension of 256, a patch size of 4, and a dropout rate of 0.1.
An Exponential Moving Average (EMA) is maintained for model weights with a decay rate of 0.999. 
All models are trained using a standard batch size of 128.
The model utilizes a SwiGLU activation function
and the final layer normalization is applied. % and step embeddings are omitted.
The bViT architecture consists of single parameter-sharing block evaluated recurrently 12 times.
On the other hand, the classical ViT has 12 independently parameterized blocks.

\section{Experiments}
Unless stated otherwise, we keep the architecture fixed and evaluate different training recipes.
In the following subsections we investigate the tradeoff between computations and model parameters, continuous time Neural ODE integration, 
and deep-supervised recurrent training.

\subsection{Architecture comparison}
\begin{figure*}[t]
    \centering
    % Left Subfigure
    \begin{subfigure}{0.32\textwidth}
        \centering
        \includegraphics[width=\linewidth]{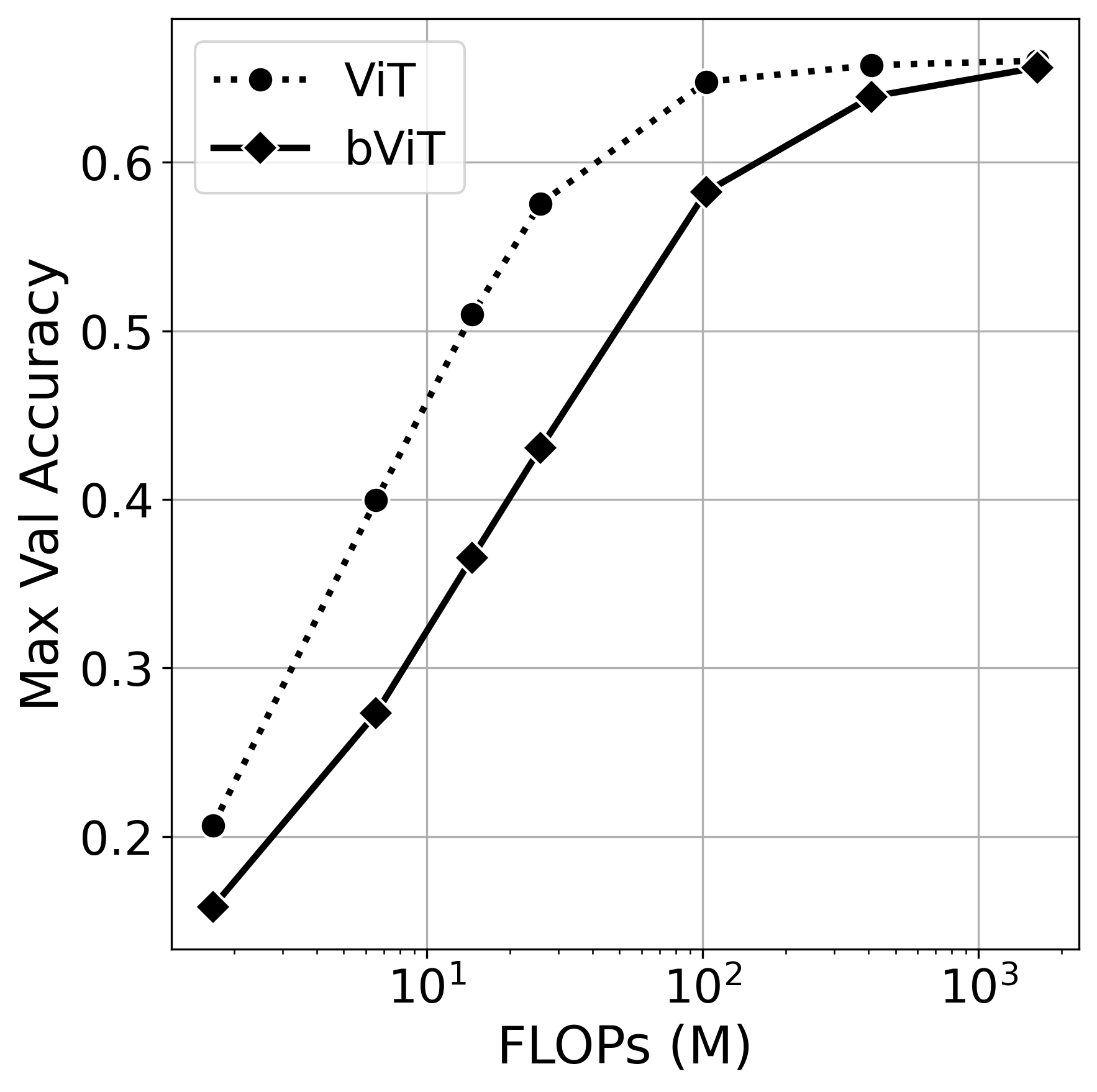}
        \caption{
        Accuracy grows as the FLOPs budget is increased. 
        The ViT outperforms the bViT. Finally, both models reach common plateau.
        }
        \label{fig:a}
    \end{subfigure}
    \hfill
    % Mid Subfigure
    \begin{subfigure}{0.32\textwidth}
        \centering
        \includegraphics[width=\linewidth]{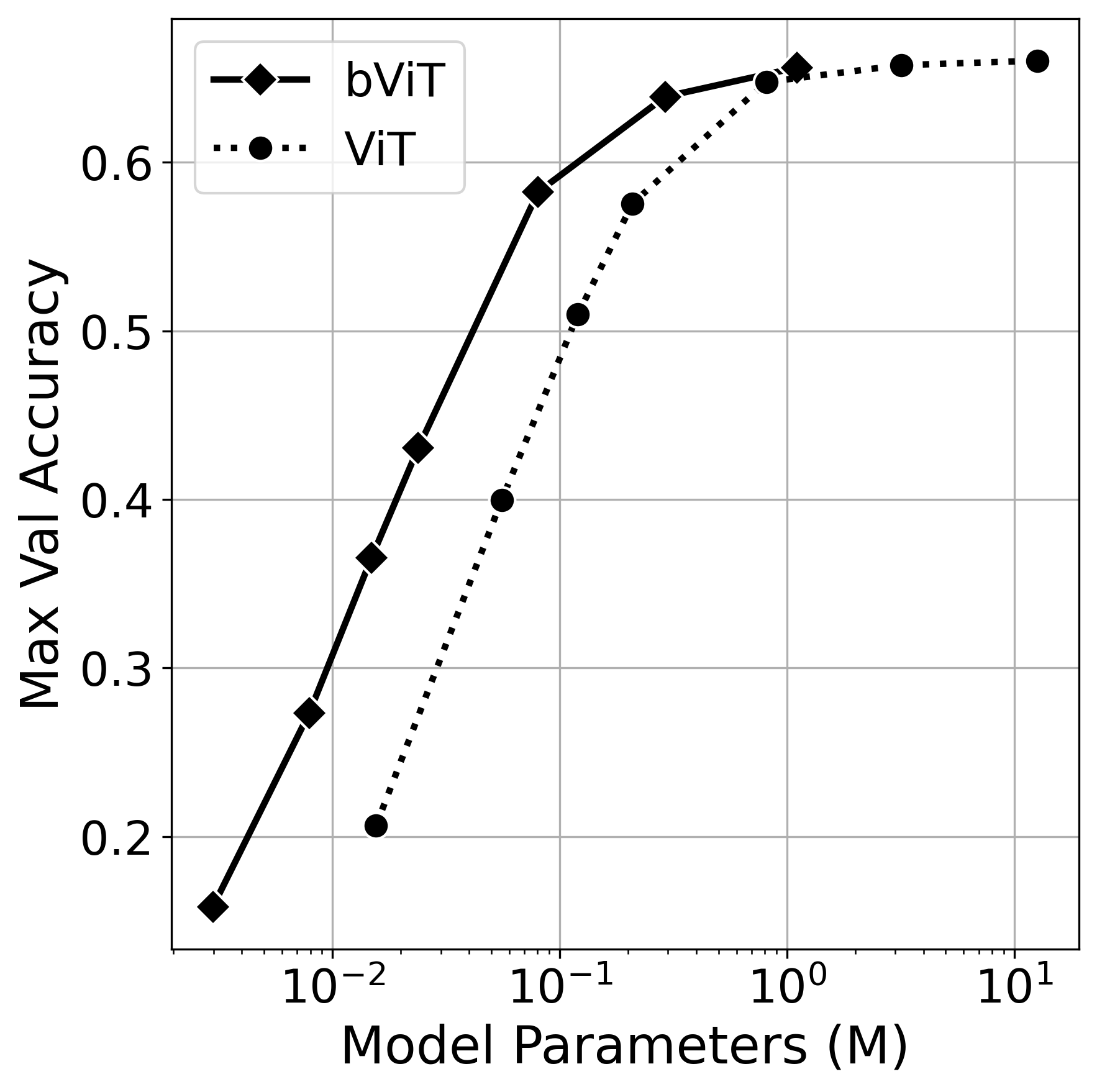}
        \caption{When the number of total network parameters is reduced, then accuracy of the bViT is better compared to ViT.}
        \label{fig:b}
    \end{subfigure}
    \hfill
    % Right Subfigure
    \begin{subfigure}{0.32\textwidth}
        \centering
        \includegraphics[width=\linewidth]{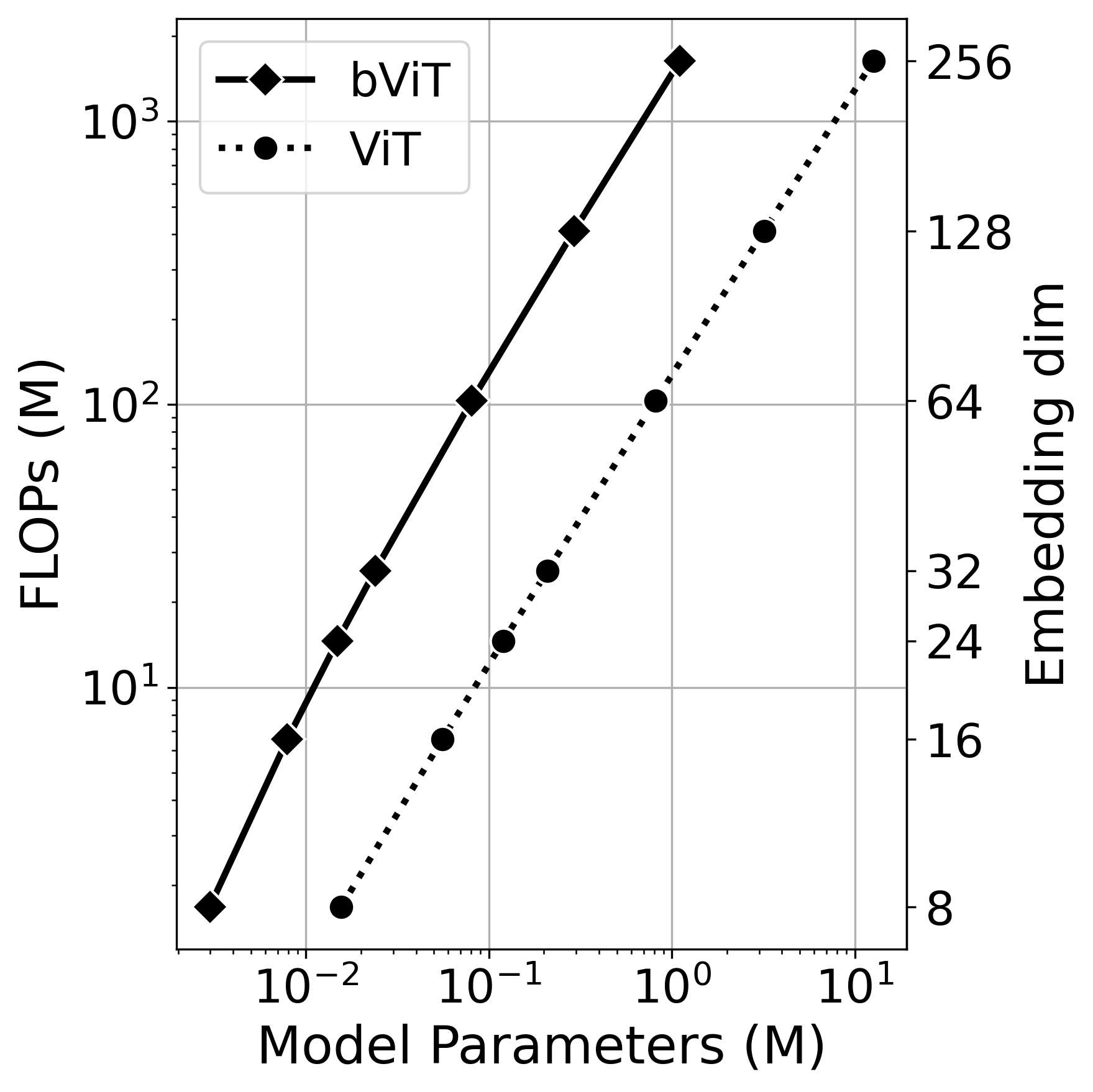}
        \caption{As the bViT shares the parameters between blocks, 
        the computational cost strictly correlates with embedding dimension.
        }
        \label{fig:c}
    \end{subfigure}
    \caption{The trade-off between the classical ViT and the recurrent bViT, evaluated across compute and scale constraints.
    }
    \label{fig:main_architecture_evaluation}
\end{figure*}

\begin{figure}%[h]
    \centering
    \includegraphics[width=0.5\textwidth]{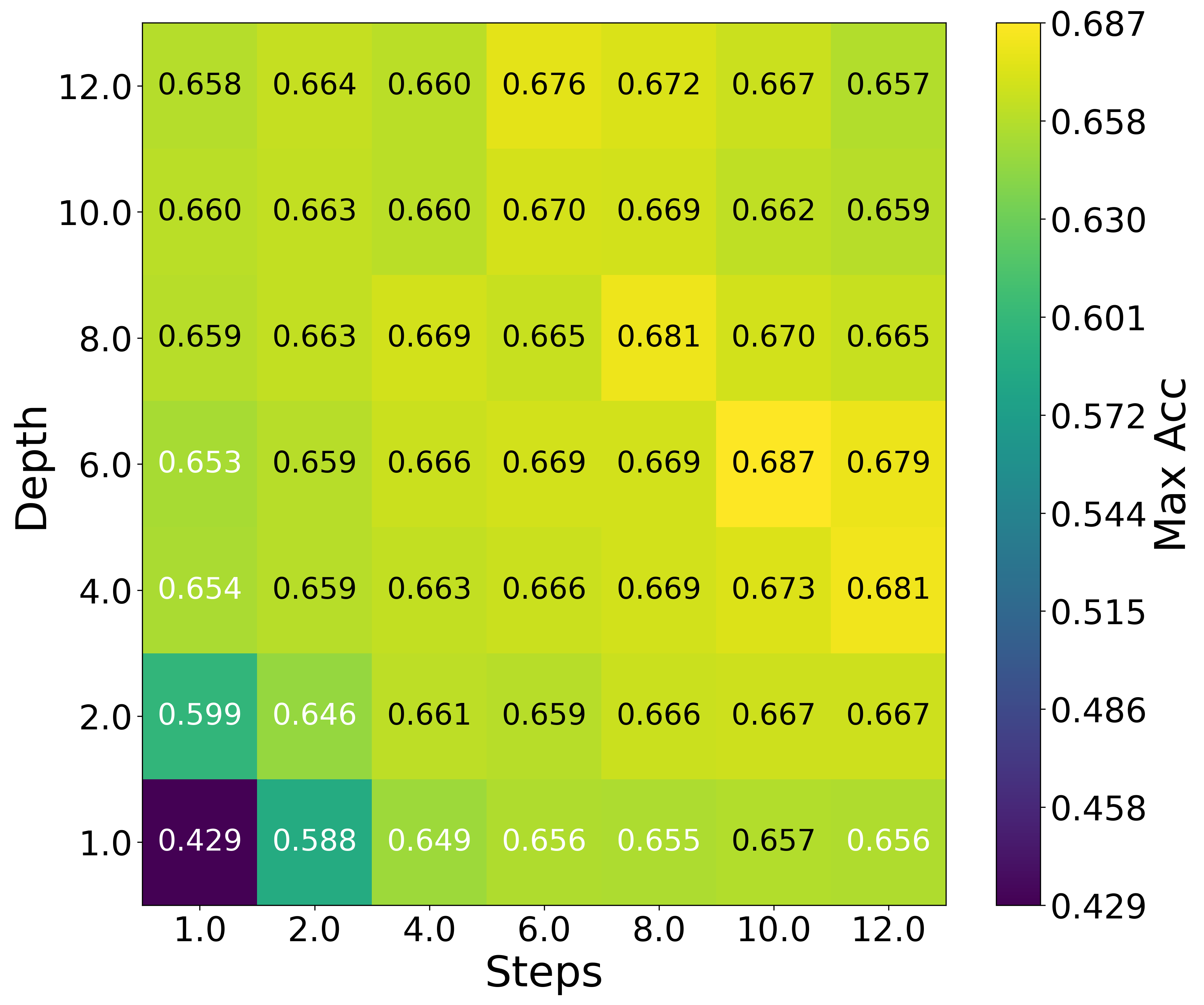}
\caption{Heat map of max accuracy. 
Sharp decline is visible when the network lacks capacity on either depth, $D$, or steps, $S$, axis.
Notice that the setup with $D=1$ and $S=12$ represent the bViT, while $D=12$ and $S=1$ is the classical ViT.
Other combinations of $D,S$ corresponds to multi-block recursive variants. 
For example model with $D=12$ and $S=12$ would have $12$ independent blocks in a stack, and the stack would be evaluated recursively $12$ times.
Such a long sequence (144 evaluations) suffers from degraded accuracy due to vanishing gradient problems.
}
\label{fig:depth_vs_steps_val_max_acc_heatmap}
\end{figure}

The comparison in Fig.~\ref{fig:main_architecture_evaluation} indicates that recurrent weight sharing in bViT induces a clear resource-dependent trade-off rather than a uniformly superior architecture. 
Under matched computational budgets, the classical ViT remains preferable, as its independently parameterized blocks provide higher accuracy for a given FLOPs budget. 
In contrast, when the dominant constraint is parameter storage or deployment memory, bViT becomes advantageous: by reusing a single transformer block across recurrent steps, it substantially reduces the parameter footprint while retaining competitive accuracy. 
Thus, standard ViTs are better suited to compute-limited regimes, whereas bViT offers a more favorable accuracy--parameter trade-off for memory-constrained deployment.
Next, we performed a detailed grid search experiment (see Fig.~\ref{fig:depth_vs_steps_val_max_acc_heatmap}) to show how 
the transformer depth, $D$, (governed by the number of unique blocks) 
and steps, $S$, (recurrent repetition of the set of block) interacts with the effective representational power.  
% Results are presented in Fig.~\ref{fig:depth_vs_steps_val_max_acc_heatmap} 
Below a critical depth or steps threshold, the shared representational space lacks the capacity to encode the required functional diversity. 
Recurrence can unroll representation space just as effectively as structural depth assuming that the number of network parameters is sufficient.
However, when the overall virtual length (steps multiplied by depth) becomes to long the model looses accuracy due to numerical discrepancies in gradient propagation.
% Consequently, increasing steps, yields diminishing returns. 
% This confirms that while recurrence efficiently unrolls computation, a minimum parameter basis is strictly required to prevent the sequence of transformations from functionally saturating.
\subsection{bViT as a Neural ODE (bViT-node) }\label{sec:bViT-node}
The bViT update $z_{n+1}=F_\theta(z_n)$ is a discrete-time dynamical
system and can be reinterpreted in continuous time~\cite{chen2018neural}.
Since a transformer block is residual, we may write
$F_\theta(z)=z+R_\theta(z)$, where $R_\theta$ denotes the non-identity
transformation induced by attention, normalization, and feed-forward
sublayers. Consistent with the standard view of residual networks as
Euler discretizations of ODEs
\cite{weinan2017proposal,Haber_2017,lu2018beyond}, %sander2022residual
the continuous-time analogue of the residual recurrence is the
\emph{state-subtracted} vector field
$\dot{z}=F_\theta(z)-z=R_\theta(z)$: explicit Euler integration with
$\Delta t=1$ gives $z_{n+1}=z_n+(F_\theta(z_n)-z_n)=F_\theta(z_n)$,
exactly recovering the discrete bViT recurrence. Although this
correspondence is well known in principle, it is easy to violate in
practice when the block is wrapped as a black-box vector field, as in
standard Neural ODE toolkits: setting $\dot{z}=F_\theta(z)$ yields,
under Euler with $\Delta t=1$,
$z_{n+1}=z_n+F_\theta(z_n)=2z_n+R_\theta(z_n)$, which accumulates the
identity component at every step. We therefore include
$\dot{z}=F_\theta(z)$ as a \emph{control condition} and quantify the
cost of this mis-specification for recurrent ViTs.

We evaluate explicit Euler, Heun2, and RK4 solvers applied to both
vector fields. \Cref{tab:EulerHeun2RK4_Swiglu} shows that the
state-subtracted formulation consistently outperforms the control by
roughly 3--6 accuracy points---a nontrivial penalty for an error that is
invisible at the API level of common ODE libraries. At matched or
comparable numbers of function evaluations (NFEs), higher-order solvers
do not uniformly improve performance despite their higher formal order.
Because the vector field is learned \emph{jointly} with the solver,
higher-order methods alter the optimization landscape by compounding
multi-stage gradient dependencies within each step; the resulting gains
should therefore be read as a solver-induced architectural bias rather
than a numerical-accuracy improvement.
Overall, explicit Euler remains a
highly competitive and robust baseline, and the optimal solver choice
depends on both the block depth and the target discretization horizon.
\begin{table}[t!]
\centering
         \caption{
         Comparison of integration schemes.
         The best and second best results for each of the integrators (Euler, Heun2, RK4) are highlighted with green and light green colors, respectively. 
         Here, $D$ and $S$ denotes the number of unique transformer blocks and the number of solver steps respectively. 
         NFE is the total number of transformer-block evaluations.
         ``NaN''\  denotes a diverged training run.
         } 
\begin{tabular}{rr|rr|rr|rr}
\toprule
& & \multicolumn{2}{c|}{Euler} & \multicolumn{2}{c|}{Heun2} & \multicolumn{2}{c}{RK4} \\
$D$ & $S$ & NFE & Acc$\uparrow$ & NFE & Acc$\uparrow$ & NFE &  Acc$\uparrow$ \\
\midrule
\multicolumn{8}{c}{NODE: $\frac{dz}{dt} = F_{\theta}(z) - z$ (state-subtracted)}  \\
\midrule
1 & 3 & 3 & 0.630 & 6 & 0.654 & 12 & 0.655 \\
1 & 6 & 6 & 0.656 & 12 & 0.657 & 24 & 0.660 \\
1 & 12 & 12 & 0.658 & 24 & 0.663 & 48 & 0.665 \\
1 & 24 & 24 & 0.667 & 48 & 0.670 & 96 & 0.671 \\
% \midrule
2 & 3 & 6 & 0.658 & 12 & 0.663 & 24 & 0.665 \\
2 & 6 & 12 & 0.664 & 24 & 0.670 & 48 & 0.668 \\
2 & 12 & 24 & 0.668 & 48 & 0.675 & 96 & 0.674 \\
2 & 24 & 48 & \cellcolor{OliveGreen!70}{0.682} & 96 & \cellcolor{SpringGreen!70}{0.689} & 192 & \cellcolor{OliveGreen!70}{0.692} \\
% \midrule
4 & 3 & 12 & 0.659 & 24 & 0.673 & 48 & 0.673 \\
4 & 6 & 24 & 0.669 & 48 & 0.676 & 96 & 0.675 \\
4 & 12 & 48 & \cellcolor{SpringGreen!70}{0.677} & 96 & \cellcolor{OliveGreen!70}{0.701} & 192 & \cellcolor{SpringGreen!70}{0.677} \\
4 & 24 & 96 & 0.669 & 192 & 0.633 & 384 & NaN \\
\midrule
\multicolumn{8}{c}{NODE: $\frac{dz}{dt} = F_{\theta}(z)$ (control)} \\
\midrule
1 & 3 & 3 & 0.591 & 6 & 0.618 & 12 & 0.637 \\
1 & 6 & 6 & 0.611 & 12 & 0.621 & 24 & 0.635 \\
1 & 12 & 12 & 0.608 & 24 & 0.619 & 48 & 0.631 \\
1 & 24 & 24 & 0.606 & 48 & 0.622 & 96 & 0.631 \\
% \midrule
2 & 3 & 6 & 0.640 & 12 & 0.646 & 24 & 0.651 \\
2 & 6 & 12 & 0.642 & 24 & 0.650 & 48 & 0.652 \\
2 & 12 & 24 & 0.642 & 48 & 0.648 & 96 & 0.654 \\
2 & 24 & 48 & 0.641 & 96 & 0.648 & 192 & 0.653 \\
% \midrule
4 & 3 & 12 & 0.658 & 24 & 0.658 & 48 & 0.664 \\
4 & 6 & 24 & 0.657 & 48 & 0.660 & 96 & \cellcolor{OliveGreen!70}{0.668} \\
4 & 12 & 48 & \cellcolor{OliveGreen!70}{0.661} & 96 & \cellcolor{SpringGreen!70}{0.663} & 192 & \cellcolor{SpringGreen!70}{0.666} \\
4 & 24 & 96 & \cellcolor{SpringGreen!70}{0.659} & 192 & \cellcolor{OliveGreen!70}{0.664} & 384 & NaN \\
\bottomrule
\end{tabular}
 \label{tab:EulerHeun2RK4_Swiglu}
\end{table}
\subsection{Deep Supervision and Extrapolation (bViT-ds)}\label{sec:bViT-ds}
Standard bViT is trained for a fixed recurrent horizon (T), and it learns a finite-horizon computation rather than necessarily learning a contractive map with a stable fixed point~\cite{byra2026bvitinvestigatingsingleblockrecurrence}. 
When evaluated for more recurrent steps than seen during training, the latent trajectory may drift away from the region in which the classifier was trained, causing extrapolation failure.

Deep Equilibrium (DEQ) models~\cite{bai2019deep,bai2020multiscale} attempt to resolve this by framing the forward pass as an implicit root-finding problem, seeking $z^\star = F_\theta(z^\star)$. 
However, in our experiments both the DEQ root-finding solvers (e.g., Anderson acceleration) and TRM-style supervision ~\cite{jolicoeur2025less} 
applied directly to bViT turned out to be divergent. 
Related supervision strategies are studied in~\cite{jolicoeur2025less,wang2025hierarchical,liao2026simple}.
In this work, we introduce bViT-ds to bridge the gap between finite-horizon reasoning and asymptotic stability.  
This bViT variant partitions the unrolled computational graph into $K$ discrete supervision stages. 
Within each supervision stage $k \in \{1, 2, 4, 8 \}$ of bViT-ds, the parameter sharing transformer block is evaluated for $S$ steps using a simple fixed point iteration, $ z_{t} = F_\theta(z_{t-1})$.
At the end of each stage, $k$, the intermediate logits $\hat{y}_k$ are computed 
and a cross-entropy penalty is applied, 
$\mathcal{L}_{\text{total}} = \frac{1}{K}\sum_{k=1}^K \mathcal{L}_{\text{CE}}(\hat{y}_k, y)$.
The $\mathcal{O}(1)$ memory scaling with respect to unrolled depth is achieved by applying state detachment after each group of $S$ steps. 
Minimizing the loss from detached intermediate states explicitly penalizes trajectory divergence and constrain the phase space required for stability.

\Cref{tab:bViT_deep_supervision_cifar100} shows that simply increasing the recurrent horizon does not monotonically improve performance. 
Accuracy initially improves with additional recurrent steps, but eventually saturates and slowly degrades.
When the same model is extrapolated beyond its training horizon, accuracy can collapse to near-random level.
Stage-wise deep supervision substantially improves extrapolation robustness---but it does not improve nominal accuracy: 
whenever the horizon is known at training time, naive extension is strictly better assuming sufficient GPU memory is available.
\begin{table}[t!]%[H]
\centering
 \caption{
  Nominal accuracy vs.\ extrapolation robustness.
 Increasing the number of steps within a single stage improves accuracy, which then plateaus and slowly degrades,
 whereas naive extrapolation beyond the training horizon collapses.
 Deep supervision (bViT-ds) helps constrain trajectory divergence and improves extrapolation robustness at the price of nominal accuracy.
 In bViT-ds we set each stage to consist of $12$ steps.
 During training, the computational graph is detached after each stage to maintain $\mathcal{O}(1)$ memory depth.
 During inference, we add additional extrapolation stages beyond the training stages.
 The number of function evaluations (overall) is denoted as NFE.
 The best and second best results are highlighted with green and light green colors, respectively.
}
\begin{tabular}{lccccc}
\toprule
              & Steps &  Supervision  & Extrapolation  &     \\
Acc$\uparrow$ & per stage &  stages       &  stages        & NFE \\
\midrule
\multicolumn{5}{c}{bViT baseline and naive extrapolation.} \\
\midrule
0.658 & 12 & 1 & 0 & 12 \\
0.029 & 12 & 1 & 4 & 60 \\
0.017 & 12 & 1 & 8 & 108 \\
\midrule
\multicolumn{5}{c}{Naive increase of the bViT loop horizon.} \\ 
\midrule
0.667 & 24 & 1 & 0 & 24 \\
\cellcolor{SpringGreen!70}{0.681} & 48 & 1 & 0 & 48 \\
\cellcolor{OliveGreen!70}{0.682} & 60 & 1 & 0 & 60 \\
0.680 & 72 & 1 & 0 & 72 \\
0.679 & 84 & 1 & 0 & 84 \\
0.676 & 96 & 1 & 0 & 96 \\
% \bottomrule
\midrule
\multicolumn{5}{c}{bViT-ds variants.} \\
\midrule
0.650 & 12 & 2 & 0 & 24 \\
0.639 & 12 & 4 & 0 & 48 \\
0.599 & 12 & 2 & 4 & 72 \\
0.620 & 12 & 4 & 4 & 96 \\
0.635 & 12 & 8 & 0 & 96 \\
0.467 & 12 & 2 & 8 & 120 \\
0.588 & 12 & 4 & 8 & 144 \\
0.630 & 12 & 8 & 4 & 144 \\
0.613 & 12 & 8 & 8 & 192 \\
\bottomrule
\end{tabular}
 \label{tab:bViT_deep_supervision_cifar100}
\end{table}
\section{Conclusion}
\label{sec:conclusions}
This work characterizes recurrent ViT training through the lens of
deployment trade-offs, continuous-time modeling, and extrapolation
stability, answering the three research questions. 
(i)~Recurrence is not a universally superior replacement for architectural depth: standard ViTs remain stronger at matched FLOPs, 
while bViTs provide a better accuracy--parameter trade-off when the model-memory footprint is the binding constraint.
(ii)~The continuous-time analogue of a residual recurrent block is the state-subtracted vector field $\dot{z}=F_\theta(z)-z$---a known correspondence whose violation we show costs 3--6 accuracy points.
Higher-order solver gains are NFE-dependent and best interpreted as solver-induced architectural bias rather than numerical-accuracy improvement. 
(iii)~Stage-wise deep supervision buys long-horizon
robustness at the price of nominal accuracy and additional computation.
% =====================================================================
% ------------------------- PAGE 5 ONLY -------------------------------
% Per the ICASSP 2027 CfP, the 5th page may contain only references,
% funding acknowledgements, and a Compliance with Ethical Standards
% statement. Keep everything below on page 5.
% ----------------------------------------------------------------------
\clearpage
{
\small
\subsection*{Funding acknowledgements}
This work was supported by Samsung AI Center, Warsaw.

\subsection*{Compliance with ethical standards}
The authors have no relevant conflicts of interest to disclose.

\bibliographystyle{IEEEbib}
\bibliography{strings}

@string{iclr={Proceedings of the International Conference on Learning Representations (ICLR)}}

@string{neurips={Advances in Neural Information Processing Systems (NeurIPS)}}

@string{aaai={Proceedings of the AAAI Conference on Artificial Intelligence (AAAI)}}

@string{icml={Proceedings of the International Conference on Machine Learning (ICML)}}

@string{springer={Springer International Publishing}}

@string{arxiv={{A}r{X}iv e-print}}

@misc{byra2026bvitinvestigatingsingleblockrecurrence,
      title={bViT: Investigating Single-Block Recurrence in Vision Transformers for Image Recognition}, 
      author={Michal Byra and Pawel Olszowiec and Grzegorz Stefanski and Grzegorz Gruszczynski and Alberto Presta},
      year={2026},
      eprint={2605.10661},
      archivePrefix={arXiv},
      primaryClass={cs.CV},
      url={https://arxiv.org/abs/2605.10661}, 
}

@inproceedings{zhang2022minivit,
  title={Minivit: Compressing vision transformers with weight multiplexing},
  author={Zhang, Jinnian and Peng, Houwen and Wu, Kan and Liu, Mengchen and Xiao, Bin and Fu, Jianlong and Yuan, Lu},
  booktitle={Proceedings of the IEEE/CVF conference on computer vision and pattern recognition},
  pages={12145--12154},
  year={2022}
}

@misc{krizhevsky2009learning,
title={Learning multiple layers of features from tiny images},
author={Krizhevsky, Alex and Hinton, Geoffrey and others},
year={2009}
}

@article{dosovitskiy2020image,
title={An image is worth 16x16 words: Transformers for image recognition at scale},
author={Dosovitskiy, Alexey and Beyer, Lucas and Kolesnikov, Alexander and Weissenborn, Dirk and Zhai, Xiaohua and Unterthiner, Thomas and Dehghani, Mostafa and Minderer, Matthias and Heigold, Georg and Gelly, Sylvain and others},
journal=iclr,
year={2020}
}

@article{Haber_2017,
   title={Stable architectures for deep neural networks},
   journal={Inverse Problems},
   publisher={IOP Publishing},
   author={Haber, Eldad and Ruthotto, Lars},
   volume={34},
   number={1},
   pages={014004},
   year={2017},
}

@article{dehghani2018universal,
  title={Universal transformers},
  author={Dehghani, Mostafa and Gouws, Stephan and Vinyals, Oriol and Uszkoreit, Jakob and Kaiser, {\L}ukasz},
  journal=iclr,
  year={2019}
}

@inproceedings{touvron2021training,
  title={Training data-efficient image transformers \& distillation through attention},
  author={Touvron, Hugo and Cord, Matthieu and Douze, Matthijs and Massa, Francisco and Sablayrolles, Alexandre and J{\'e}gou, Herv{\'e}},
  booktitle={International conference on machine learning},
  pages={10347--10357},
  year={2021},
  organization={PMLR}
}

@article{wang2025hierarchical,
  title={Hierarchical reasoning model},
  author={Wang, Guan and Li, Jin and Sun, Yuhao and Chen, Xing and Liu, Changling and Wu, Yue and Lu, Meng and Song, Sen and Yadkori, Yasin Abbasi},
  journal={arXiv preprint arXiv:2506.21734},
  year={2025}
}

@article{liao2026simple,
  title={Simple Recursive Model: Simplified, Single-State Reasoning with Skip Connections},
  author={Liao, Qianli and Poggio, Tomaso},
  journal={Preprint},
  year={2026}
}

@article{bai2020multiscale,
  title={Multiscale deep equilibrium models},
  author={Bai, Shaojie and Koltun, Vladlen and Kolter, J Zico},
  journal={Advances in neural information processing systems},
  volume={33},
  pages={5238--5250},
  year={2020}
}

@inproceedings{sun2025transformer,
  title={Transformer layers as painters},
  author={Sun, Qi and Pickett, Marc and Nain, Aakash Kumar and Jones, Llion},
  booktitle={Proceedings of the AAAI Conference on Artificial Intelligence},
  volume={39},
  number={24},
  pages={25219--25227},
  year={2025}
}

@article{jacobs2025block,
  title={Block-Recurrent Dynamics in Vision Transformers},
  author={Jacobs, Mozes and Fel, Thomas and Hakim, Richard and Brondetta, Alessandra and Ba, Demba and Keller, T Andy},
  journal={arXiv preprint arXiv:2512.19941},
  year={2025}
}

@article{saunshi2025reasoning,
title={Reasoning with latent thoughts: On the power of looped transformers},
author={Saunshi, Nikunj and Dikkala, Nishanth and Li, Zhiyuan and Kumar, Sanjiv and Reddi, Sashank J},
journal=iclr,
year={2025}
}

@article{bai2019deep,
  title={Deep equilibrium models},
  author={Bai, Shaojie and Kolter, J Zico and Koltun, Vladlen},
  journal=neurips,
  volume={32},
  year={2019}
}

@article{chen2018neural,
  title={Neural ordinary differential equations},
  author={Chen, Ricky TQ and Rubanova, Yulia and Bettencourt, Jesse and Duvenaud, David K},
  journal={Advances in neural information processing systems},
  volume={31},
  year={2018}
}

@article{jolicoeur2025less,
  title={Less is more: Recursive reasoning with tiny networks},
  author={Jolicoeur-Martineau, Alexia},
  journal={arXiv preprint arXiv:2510.04871},
  year={2025}
}

@inproceedings{garg2025revealing,
  title={Revealing the utilized rank of subspaces of learning in neural networks},
  author={Garg, Isha and Koguchi, Christian and Verma, Eshan and Ulbricht, Daniel},
  booktitle={Proceedings of the AAAI Symposium Series},
  volume={5},
  number={1},
  pages={151--158},
  year={2025}
}

@article{akiba2025evolutionary,
  title={Evolutionary optimization of model merging recipes},
  author={Akiba, Takuya and Shing, Makoto and Tang, Yujin and Sun, Qi and Ha, David},
  journal={Nature Machine Intelligence},
  volume={7},
  number={2},
  pages={195--204},
  year={2025},
  publisher={Nature Publishing Group UK London}
}

@inproceedings{he2022masked,
  title={Masked autoencoders are scalable vision learners},
  author={He, Kaiming and Chen, Xinlei and Xie, Saining and Li, Yanghao and Doll{\'a}r, Piotr and Girshick, Ross},
  booktitle={Proceedings of the IEEE/CVF conference on computer vision and pattern recognition},
  pages={16000--16009},
  year={2022}
}

@inproceedings{caron2021emerging,
  title={Emerging properties in self-supervised vision transformers},
  author={Caron, Mathilde and Touvron, Hugo and Misra, Ishan and J{\'e}gou, Herv{\'e} and Mairal, Julien and Bojanowski, Piotr and Joulin, Armand},
  booktitle={Proceedings of the IEEE/CVF international conference on computer vision},
  pages={9650--9660},
  year={2021}
}

@article{lan2019albert,
  title={Albert: A lite bert for self-supervised learning of language representations},
  author={Lan, Zhenzhong and Chen, Mingda and Goodman, Sebastian and Gimpel, Kevin and Sharma, Piyush and Soricut, Radu},
  journal=iclr,
  year={2020}
}

@article{loshchilov2017decoupled,
  title={Decoupled weight decay regularization},
  author={Loshchilov, Ilya and Hutter, Frank},
  journal=iclr,
  year={2019}
}

@inproceedings{shen2022sliced,
  title={Sliced recursive transformer},
  author={Shen, Zhiqiang and Liu, Zechun and Xing, Eric},
  booktitle={European Conference on Computer Vision},
  pages={727--744},
  year={2022},
  organization={Springer}
}

@article{weinan2017proposal,
  title={A proposal on machine learning via dynamical systems},
  author={E, Weinan},
  journal={Communications in Mathematics and Statistics},
  volume={5},
  number={1},
  pages={1--11},
  year={2017}
}

@inproceedings{lu2018beyond,
  title={Beyond finite layer neural networks: Bridging deep architectures and numerical differential equations},
  author={Lu, Yiping and Zhong, Aoxiao and Li, Quanzheng and Dong, Bin},
  booktitle=icml,
  pages={3276--3285},
  year={2018}
}
}
\end{document}